\documentclass[11pt,a4paper]{article}
\usepackage[hyperref]{acl2019}
\usepackage{times}
\usepackage{latexsym}
\usepackage{amstext}
\usepackage{mathtools}
\usepackage{amsmath}
\usepackage{bbm}
\usepackage{bbold}
\usepackage{forest}
\usepackage{qtree}
\usepackage{url}
\usepackage{multirow}
\usepackage{graphicx} %package to manage images
\graphicspath{ {images/} }
\usepackage{tabularx}
\usepackage{ragged2e}
\usepackage{tkz-berge}
\usepackage{tikz}
\usetikzlibrary{matrix,chains,positioning,decorations,snakes,shapes.arrows}
\usepackage{pgfplots}
\usepackage{subcaption}
\usepackage{array}
\usepackage{enumitem}
\usepackage{comment}
\newcolumntype{P}[1]{>{\centering\arraybackslash}p{#1}}
\usepackage{amsthm}
\theoremstyle{definition}

\aclfinalcopy % Uncomment this line for the final submission
\def\aclpaperid{387} %  Enter the acl Paper ID here

\newcounter{Lcount}
\newcommand{\squishenum}{
	\begin{list}{\arabic{Lcount}. }
		{ \usecounter{Lcount}
			\setlength{\itemsep}{0pt}
			\setlength{\parsep}{0pt}
			\setlength{\topsep}{0pt}
			\setlength{\partopsep}{0pt}
			\setlength{\leftmargin}{2em}
			\setlength{\labelwidth}{1.5em}
			\setlength{\labelsep}{0.5em} } }
	
	\newcommand{\squishletter}{
		\begin{list}{\alph{Lcount}. }
			{ \usecounter{Lcount}
				\setlength{\itemsep}{0pt}
				\setlength{\parsep}{0pt}
				\setlength{\topsep}{0pt}
				\setlength{\partopsep}{0pt}
				\setlength{\leftmargin}{2em}
				\setlength{\labelwidth}{1.5em}
				\setlength{\labelsep}{0.5em} } }
		
		\newcommand{\squishlist}{
			\begin{list}{$\bullet$}
				{ \usecounter{Lcount}
					\setlength{\itemsep}{0pt}
					\setlength{\parsep}{0pt}
					\setlength{\topsep}{0pt}
					\setlength{\partopsep}{0pt}
					\setlength{\leftmargin}{2em}
					\setlength{\labelwidth}{1.5em}
					\setlength{\labelsep}{0.5em} } }

			\newcommand{\squishend}{
		\end{list} }
		\usepackage{pgf}
		\usetikzlibrary{arrows,decorations.pathmorphing,backgrounds,positioning,fit,petri,shapes.misc, arrows.meta,shapes.geometric,decorations.markings,calc,shadows.blur}
		\definecolor{myblue}{RGB}{0,128,255}
		\definecolor{myorange}{RGB}{211, 84, 0}
		\definecolor{mypurple}{RGB}{142, 68, 173}
		\definecolor{mygrey}{RGB}{158, 158, 158}
		\definecolor{lowpurple}{RGB}{204,153,255}
		\definecolor{lowwhite}{RGB}{255,255,255}
		\definecolor{verylowpurple}{RGB}{255,102,102}
		\definecolor{embcolor}{RGB}{255,255,255}
		\definecolor{myred}{RGB}{231, 76, 60}
		\definecolor{mygreen}{RGB}{162, 217, 206} 
		\definecolor{fontgrey}{RGB}{44, 62, 80}
		\definecolor{lowpurple}{RGB}{210, 180, 222}
		\definecolor{mypumpkin}{RGB}{229, 152, 102}
		\definecolor{lowgreen}{RGB}{171, 235, 198}
		\definecolor{lowred}{RGB}{245, 183, 177}
		\definecolor{pink}{RGB}{252,146,114}

\title{\texttt{Quantity Tagger}: A Latent-Variable Sequence Labeling Approach to Solving Addition-Subtraction Word Problems}

\author{%
	Yanyan Zou \and Wei Lu\\
	StatNLP Research Group\\
	Singapore University of Technology and Design \\
	{\tt yanyan\_zou@mymail.sutd.edu.sg, luwei@sutd.edu.sg} \\
}

\date{}

\begin{document}
\maketitle
\begin{abstract}
An arithmetic word problem typically includes a textual description containing several constant quantities.
The key to solving the problem is to reveal the underlying mathematical relations (such as addition and subtraction) among quantities, and then generate equations to find solutions.
This work presents a novel approach, \emph{Quantity Tagger}, that automatically discovers such hidden relations by tagging each quantity with a \emph{sign} corresponding to one type of mathematical operation.
For each quantity, we assume there exists a latent, variable-sized {\em quantity span} surrounding the quantity token in the text, which conveys information useful for determining its sign.
Empirical results show that our method achieves 5 and 8 points of accuracy gains on two datasets respectively, compared to prior approaches. 
\end{abstract}

\section{Introduction}
Teaching machines to automatically solve arithmetic word problems, exemplified by two problems in Figure \ref{fig:problem_definition}, is a long-standing Artificial Intelligence (AI) task \cite{bobrow1964question,mukherjee2008review}.
Recent research \cite{hosseini2014learning,kushman2014learning,roy2015solving,wang2017deep,wang2018mathdqn,wang2018translating} focused on designing algorithms to automatically solve arithmetic word problems.
One line of prior works designed rules  \cite{mukherjee2008review,hosseini2014learning} or templates \cite{kushman2014learning,zhou2015learn,mitra2016learning} to map problems to expressions, where rules or templates are collected from training data.

\begin{figure}[tp]
	\scalebox{0.68}{
		\centering
		\begin{tabular}{|p{10.2cm}|}
			\hline 
			\textbf{Problem 1}: {\em A worker at a medical lab is studying blood samples. $\mathbf{2}$ samples contained a total of $\mathbf{7341}$ blood cells. The first sample contained $\mathbf{4221}$ blood cells. How many blood cells were in the second sample?}\\
			\hline
			\textbf{Prediction}: ${{\color{red}{({\mathbb{0})}}\times}}\mathbf{2}+{\color{red}({\mathbb{+1}})\times}\mathbf{7341}+{\color{red}({{\mathbb{-1}}})\times}\mathbf{4221}+{\color{red}({{\mathbb{-1}}})\times} x=0$ \\
			\textbf{Equation}: $7341-4221-x=0$ \\
			\textbf{Solution}: $x=3120$  \\
			\hline
			\hline     
			\textbf{Problem 2}: {\em There are $\mathbf{22}$ walnut trees currently in the park. Park workers will plant walnut trees today. When the workers are finished there will be $\mathbf{55}$ walnut trees in the park. How many walnut trees did the workers plant today?}\\
			\hline
			\textbf{Prediction}: ${{\color{red}{\mathbb{(+1)}}\times}}\mathbf{22} + {{\color{red}{\mathbb{(-1)}}\times}}\mathbf{55} +  {{\color{red}{\mathbb{(+1)}}\times}}x = 0$ \\       
			\textbf{Equation}: $22 - 55 +x=0$ \\
			\textbf{Solution}: $x=33$      \\           
			\hline        
		\end{tabular}
	}
	\vspace{-2mm}
	\caption{Two examples of arithmetic word problems described in English  with answers.}
	\label{fig:problem_definition}
	\vspace{-5mm}
\end{figure}

However, it would be non-trivial and expensive to acquire a general set of rules or templates.
Furthermore, such approaches typically require additional annotations.
The {\em addition-subtraction} problems, which constitute the most fundamental class of arithmetic word problems, have been the focus for many previous works \cite{hosseini2014learning,mitra2016learning}.
We also focus on this important task in this work.
Our key observation is that essentially solving such a class of problems can be tackled from a sequence labeling perspective.
This motivates us to build a novel sequence labeling approach, namely \emph{Quantity Tagger}.
The approach tags each quantity in the text with a label that indicates a specific mathematical operation.

Taking Problem 1 from Figure \ref{fig:problem_definition} as an example,
three constant quantities ``${2}$'',``${7341}$'' and ``${4221}$'' sequentially appear in the problem text.
We further introduce an unknown quantity $x$ corresponding to the question sentence.
From the problem description, one can form an equation ``$7341-4221-x=0$",  based on which we can obtain the solution to $x$.
This equation is mathematically equivalent to
``${\mathbb{0}\times}2+{(\mathbb{+1})\times}7341+{(\mathbb{-1})\times}4221+{\mathbb{(-1)}\times}x=0$'' where ``$\mathbb{0},\mathbb{+1},\mathbb{-1},\mathbb{-1}$'' are {\em sign}s associated with the quantities ``$2,7341,4221,x$''.

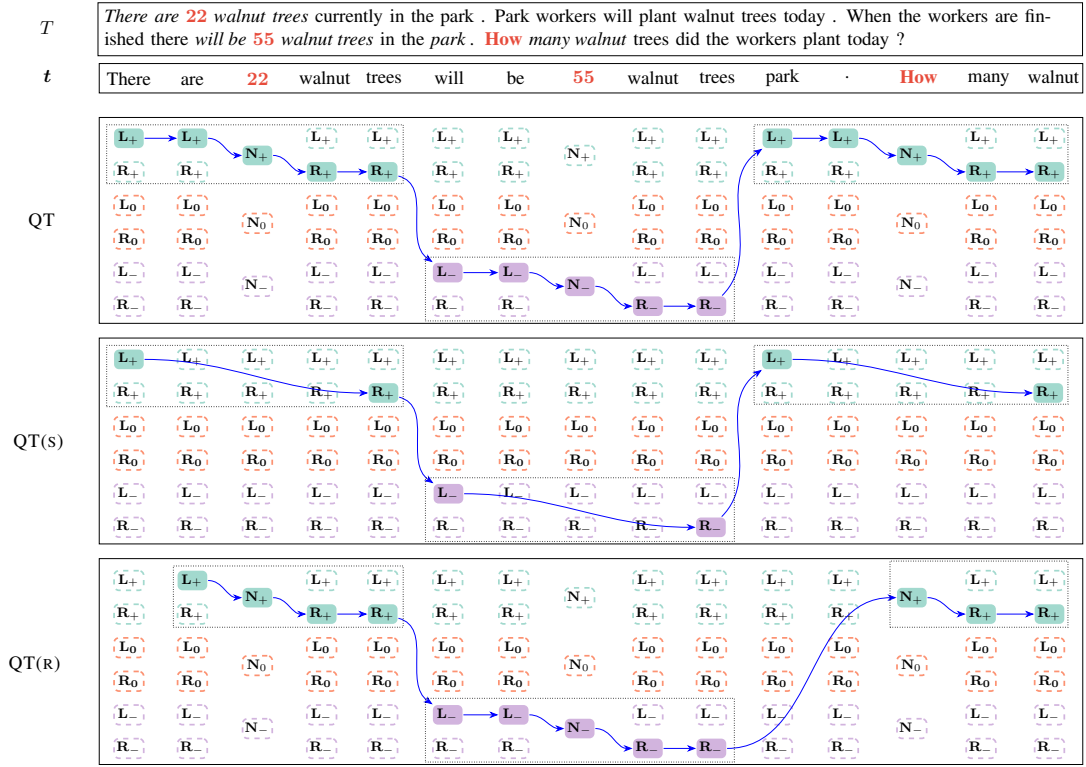
\begin{figure*}[t!]
	\centering
	\scalebox{0.65}{
		\begin{tikzpicture}[
		node distance=2.5mm and 4.0mm, >=Stealth, 
		word/.style={draw=none, minimum height=5mm, rectangle, inner sep=0pt},
		% R nodes
		%		oblabel/.style={draw=none, circle, minimum height=6mm, minimum width=6mm,line width=1pt, inner sep=2pt, fill=gray, minimum height=4mm, rectangle, rounded corners=1mm,text=fontgrey, label={center:\textsubscript{$\mathbf{R_0}$}}},
		%		o0blabel/.style={draw=none, circle, minimum height=6mm, minimum width=6mm,line width=1pt, inner sep=2pt, minimum height=4mm, rectangle, rounded corners=1mm,fill=gray, text=black, label={center:\textsubscript{$\mathbf{R_-}$}}},
		%		o1blabel/.style={draw=none, circle, minimum height=6mm, minimum width=6mm,line width=1pt, inner sep=2pt,minimum height=4mm, rectangle, rounded corners=1mm, fill=gray, text=black, label={center:\textsubscript{$\mathbf{R_+}$}}},
		rpblabel/.style={draw=mygreen,dashed, minimum height=6mm, minimum width=6mm,line width=1pt, inner sep=2pt,minimum height=4mm, rectangle, rounded corners=1mm, fill=white, text=black, label={center:\textsubscript{$\mathbf{R_+}$}}},
		r0label/.style={draw=pink,dashed, minimum height=6mm, minimum width=6mm,line width=1pt, inner sep=2pt, fill=white, minimum height=4mm, rectangle, rounded corners=1mm,text=fontgrey, label={center:\textsubscript{$\mathbf{R_0}$}}},
		rnlabel/.style={draw=lowpurple,dashed, minimum height=6mm, minimum width=6mm,line width=1pt, inner sep=2pt, minimum height=4mm, rectangle, rounded corners=1mm,fill=white, text=black, label={center:\textsubscript{$\mathbf{R_-}$}}},
		srpblabel/.style={draw=none, minimum height=6mm, minimum width=6mm,line width=1pt, inner sep=2pt,minimum height=4mm, rectangle, rounded corners=1mm, fill=mygreen, text=black, label={center:\textsubscript{$\mathbf{R_+}$}}},
		sr0label/.style={draw=none, minimum height=6mm, minimum width=6mm,line width=1pt, inner sep=2pt, fill=pink, minimum height=4mm, rectangle, rounded corners=1mm,text=fontgrey, label={center:\textsubscript{$\mathbf{R_0}$}}},
		srnlabel/.style={draw=none, minimum height=6mm, minimum width=6mm,line width=1pt, inner sep=2pt, minimum height=4mm, rectangle, rounded corners=1mm,fill=lowpurple, text=black, label={center:\textsubscript{$\mathbf{R_-}$}}},
		% N nodes
		n0label/.style={draw=pink,dashed, minimum height=6mm, minimum width=6mm,line width=1pt, inner sep=2pt, minimum height=4mm, rectangle, rounded corners=1mm,fill=white, text=fontgrey, label={center:\textsubscript{$\mathbf{N}_{0}$}}},
		nnlabel/.style={draw=lowpurple,dashed, minimum height=6mm, minimum width=6mm,line width=1pt, inner sep=2pt, minimum height=4mm, rectangle, rounded corners=1mm,fill=white, text=black, label={center: \textsubscript{$\mathbf{N_-}$}}},
		nplabel/.style={draw=mygreen,dashed, minimum height=6mm, minimum width=6mm,line width=1pt, inner sep=2pt, minimum height=4mm, rectangle, rounded corners=1mm,fill=white, text=black, label={center:\textsubscript{$\mathbf{N_+}$}}},
		sn0label/.style={draw=none, minimum height=6mm, minimum width=6mm,line width=1pt, inner sep=2pt, minimum height=4mm, rectangle, rounded corners=1mm,fill=pink, text=fontgrey, label={center:\textsubscript{$\mathbf{N}_{0}$}}},
		snnlabel/.style={draw=none, minimum height=6mm, minimum width=6mm,line width=1pt, inner sep=2pt, minimum height=4mm, rectangle, rounded corners=1mm,fill=lowpurple, text=black, label={center: \textsubscript{$\mathbf{N_-}$}}},
		snplabel/.style={draw=none, minimum height=6mm, minimum width=6mm,line width=1pt, inner sep=2pt, minimum height=4mm, rectangle, rounded corners=1mm,fill=mygreen, text=black, label={center:\textsubscript{$\mathbf{N_+}$}}},
		% L Nodes
		lnlabel/.style={draw=lowpurple,dashed, minimum height=6mm, minimum width=6mm,line width=1pt, inner sep=2pt, minimum height=4mm, rectangle, rounded corners=1mm,fill=white, text=black, label={center:\textsubscript{$\mathbf{L_-}$}}},
		lplabel/.style={draw=mygreen,dashed, minimum height=6mm, minimum width=6mm,line width=1pt, inner sep=2pt, minimum height=4mm, rectangle, rounded corners=1mm, fill=white, text=fontgrey, label={center:\textsubscript{$\mathbf{L}_{+}$}}},
		l0label/.style={draw=pink,dashed, minimum height=6mm, minimum width=6mm,line width=1pt, inner sep=2pt, minimum height=4mm, rectangle, rounded corners=1mm,fill=white, text=black, label={center:\textsubscript{$\mathbf{L_0}$}}},
		slnlabel/.style={draw=none, minimum height=6mm, minimum width=6mm,line width=1pt, inner sep=2pt, minimum height=4mm, rectangle, rounded corners=1mm,fill=lowpurple, text=black, label={center:\textsubscript{$\mathbf{L_-}$}}},
		slplabel/.style={draw=none, minimum height=6mm, minimum width=6mm,line width=1pt, inner sep=2pt, minimum height=4mm, rectangle, rounded corners=1mm, fill=mygreen, text=fontgrey, label={center:\textsubscript{$\mathbf{L}_{+}$}}},
		sl0label/.style={draw=none, minimum height=6mm, minimum width=6mm,line width=1pt, inner sep=2pt, minimum height=4mm, rectangle, rounded corners=1mm,fill=pink, text=black, label={center:\textsubscript{$\mathbf{L_0}$}}},
		noalabel/.style={draw=none, circle, minimum height=6mm, minimum width=6mm,line width=1pt, inner sep=2pt,minimum height=4mm, rectangle, rounded corners=1mm, fill=white, text=white, label={}},
		chainLine/.style={line width=1pt,-, color=black}
		]
		
		noalabel/.style={draw=none, circle, minimum height=6mm, minimum width=6mm,line width=1pt, inner sep=2pt,minimum height=4mm, rectangle, rounded corners=1mm, fill=white, text=white, label={}},
		chainLine/.style={line width=1pt,-, color=black}
		]
		\node [word] (at) {\textcolor{black}{There}};
		\node [word, left = of at, xshift=2mm,yshift=10mm] (idx11) {$T$\textcolor{white}{space}};
		\node [word, left = of at, xshift=1.5mm] (idx1) {$\boldsymbol{t}$\textcolor{white}{space}};
		\node [word, right= of at, xshift=1.9mm,yshift=-0.5mm] (are) {are};
		\node [word, right= of are, xshift=4.5mm, yshift=0.4mm] (22) {\textbf{\textcolor{myred}{$\mathbf{22}$}}};
		\node [word, right= of 22,xshift=2mm ,yshift=0.3mm] (comma)  {walnut};
		\node [word, right= of comma,xshift=-1mm, yshift=0.0mm] (we)  {trees};
		\node [word, right= of we,xshift=2.4mm, yshift=0.0mm] (gave) {will};
		\node [word, right= of gave,xshift=4.5mm] (an)  {be};
		\node [word, right= of an,xshift=5.6mm, yshift=0.0mm] (account) {\textbf{\textcolor{myred}{$\mathbf{55}$}}};
		\node [word, right= of account,xshift=2.5mm, yshift=0.0mm] (of) {\textcolor{black}{walnut}};
		\node [word, right= of of,xshift=0.0mm] (the2) {\textcolor{black}{trees}};
		\node [word, right= of the2, xshift=2mm, yshift=-0.0mm] (current) {\textcolor{black}{park}};
		\node [word, right= of current, xshift=5mm, yshift=0.0mm] (education) {\textcolor{black}{.}};
		\node [word, right= of education, xshift=6.0mm, yshift=-0.0mm] (system) {\textbf{\textcolor{myred}{{How}}}};
		\node [word, right= of system, xshift=2mm, yshift=-0.6mm] (direction) {\textcolor{black}{{many}}};
		\node [word, right= of direction, xshift=-0.5mm, yshift=+0.8mm] (they) {\textcolor{black}{{walnut}}};
		%		\node [word, right= of they, xshift=3mm, yshift=-0.0mm] (stop) {\textcolor{myred}{came}};
		%		\node [word, right= of stop, xshift=6mm, yshift=-0.0mm] (stop) {.};
		
		%	\draw [line width=0.8pt,->, dashed, color=mygrey]  (seminar)  to [out=120,in=50, looseness=1] (at);
		%	\draw [line width=0.8pt,->, dashed, color=mygrey]  (seminar)  to [out=120,in=50, looseness=1] (the1); 
		%	\draw [line width=0.8pt,->, color=myblue]  (gave)  to [out=120,in=60, looseness=0.8] (seminar) ; % 
		%	\draw [line width=0.8pt,->, dashed, color=mygrey]  (gave)  to [out=120,in=60, looseness=1.05] (we) ;  %% marked to was
		%	\draw [line width=0.8pt,->, dashed, color=mygrey]  (gave)  to [out=120,in=60, looseness=1.05] (comma) ;  %% marked to was
		%	\draw [line width=0.8pt,->, dashed, color=mygrey]  (account)  to [out=120,in=60, looseness=1] (an) ; %% markked to empty
		%	\draw [line width=0.8pt,->, color=myblue]  (gave)  to [out=60,in=120, looseness=0.9] (account) ; %% marked to even
		%	\draw [line width=0.8pt,->, dashed, color=mygrey]  (system)  to [out=130,in=50, looseness=0.6] (of) ; 
		%	\draw [line width=0.8pt,->, dashed, color=mygrey]  (system)  to [out=130,in=50, looseness=0.6]  (the2) ;
		%	\draw [line width=0.8pt,->, dashed, color=mygrey]  (system)  to [out=130,in=50, looseness=0.6]  (current) ;
		%	\draw [line width=0.8pt,->, dashed, color=mygrey]  (system)  to [out=130,in=50, looseness=0.6]  (education) ;
		%	\draw [line width=0.8pt,->, color=myblue]  (account)  to [out=50,in=130, looseness=0.6] (system)  ; %% eveb to ink

		%QT
		%L+ nodes
		\node [slplabel](lp0) [below= of at, yshift=-5mm] {}; %%The
		\node [word, left = of lp0, xshift=-2.8mm,yshift=-17mm] {{\textsc{QT\textcolor{white}{(s)}}}}; %{\textbf{QT}};
		\node [slplabel](lp1) [right=of lp0, xshift=2.5mm] {}; 
		\node [snplabel](np1) [right=of lp1, xshift=2.8mm, yshift=-3.5mm] {}; 
		\node [lplabel](lp3) [right=of lp1, xshift=15.8mm] {}; 
		\node [lplabel](lp4) [right=of lp3,xshift=2.0mm] {}; 
		\node [lplabel](lp5) [right=of lp4,xshift=2.8mm] {}; 
		\node [lplabel](lp6) [right=of lp5, xshift=3mm] {}; %% even
		\node [nplabel](np2) [right=of lp6, xshift=3mm, yshift=-3.5mm] {}; %% with
		\node [lplabel](lp8) [right=of lp6, xshift=16.8mm] {}; 
		\node [lplabel](lp9) [right=of lp8, xshift=2.5mm] {}; 
		\node [slplabel](lp10) [right=of lp9, xshift=3mm] {}; 
		\node [slplabel](lp11) [right=of lp10, xshift=3mm] {}; 
		\node [snplabel](np3) [right=of lp11,xshift=3.5mm,yshift=-3.5mm] {}; 
		\node [lplabel](lp13) [right=of lp11,xshift=17.5mm] {}; 
		\node [lplabel](lp14) [right=of lp13,xshift=3.5mm] {}; 
		%R+ nodes
		\node [rpblabel](rp0) [below= of lp0] {}; %%The
		\node [word, left = of rp0, xshift=-2.4mm] {}; %{\textbf{\textsc{QT(s)}}};
		\node [rpblabel](rp1) [below=of lp1] {}; %%cartridge
		%		\node [rpblabel](rp2) [below=of lp2] {}; %% was
		\node [srpblabel](rp3) [below=of lp3] {}; %% marked
		\node [srpblabel](rp4) [below=of lp4] {}; %% as 
		\node [rpblabel](rp5) [below=of lp5] {}; %% empty
		\node [rpblabel](rp6) [below=of lp6] {}; %% even
		%		\node [rpblabel](rp7) [below=of lp7] {}; %% with
		\node [rpblabel](rp8) [below=of lp8] {}; 
		\node [rpblabel](rp9) [below=of lp9] {}; 
		\node [rpblabel](rp10) [below=of lp10] {}; 
		\node [rpblabel](rp11) [below=of lp11] {}; 
		%		\node [rpblabel](rp12) [below=of lp12] {}; 
		\node [srpblabel](rp13) [below=of lp13] {}; 
		\node [srpblabel](rp14) [below=of lp14] {}; 
		
		%L0 nodes
		\node [l0label](l00) [below= of rp0] {}; %%The
		\node [word, left = of l00, xshift=-3.4mm]{};
		\node [l0label](l01) [below=of rp1] {}; %%cartridge
		\node [n0label](n01) [below=of np1, yshift=-7.0mm] {}; %% was
		\node [l0label](l03) [below=of rp3] {}; %% marked
		\node [l0label](l04) [below=of rp4] {}; %% as 
		\node [l0label](l05) [below=of rp5] {}; %% empty
		\node [l0label](l06) [below=of rp6] {}; %% even
		\node [n0label](n02) [below=of np2, yshift=-7.0mm] {}; %% with
		\node [l0label](l08) [below=of rp8] {}; 
		\node [l0label](l09) [below=of rp9] {}; 
		\node [l0label](l010) [below=of rp10] {}; 
		\node [l0label](l011) [below=of rp11] {}; 
		\node [n0label](n03) [below=of np3, yshift=-7.0mm] {}; 
		\node [l0label](l013) [below=of rp13] {}; 
		\node [l0label](l014) [below=of rp14] {}; 
		%R0 nodes
		\node [r0label](r00) [below= of l00] {}; %%The
		\node [word, left = of r00, xshift=-1.4mm]{};
		\node [r0label](r01) [below=of l01] {}; %%cartridge
		%		\node [r0label](r02) [below=of l02] {}; %% was
		\node [r0label](r03) [below=of l03] {}; %% marked
		\node [r0label](r04) [below=of l04] {}; %% as 
		\node [r0label](r05) [below=of l05] {}; %% empty
		\node [r0label](r06) [below=of l06] {}; %% even
		%		\node [r0label](r07) [below=of l07] {}; %% with
		\node [r0label](r08) [below=of l08] {}; 
		\node [r0label](r09) [below=of l09] {}; 
		\node [r0label](r010) [below=of l010] {}; 
		\node [r0label](r011) [below=of l011] {}; 
		%		\node [r0label](r012) [below=of l012] {}; 
		\node [r0label](r013) [below=of l013] {}; 
		\node [r0label](r014) [below=of l014] {}; 
		
		%L- nodes 
		\node [lnlabel](ln0) [below= of r00] {}; %%The
		\node [word, left = of ln0, xshift=-1.4mm]{} ;
		\node [lnlabel](ln1) [below=of r01] {}; %%cartridge
		\node [nnlabel](nn1) [below=of n01,yshift=-6mm] {}; %% was
		\node [lnlabel](ln3) [below=of r03] {}; %% marked
		\node [lnlabel](ln4) [below=of r04] {}; %% as 
		\node [slnlabel](ln5) [below=of r05] {}; %% empty
		\node [slnlabel](ln6) [below=of r06] {}; %% even
		\node [snnlabel](nn2) [below=of n02,yshift=-6mm] {}; %% with
		\node [lnlabel](ln8) [below=of r08] {}; 
		\node [lnlabel](ln9) [below=of r09] {}; 
		\node [lnlabel](ln10) [below=of r010] {}; 
		\node [lnlabel](ln11) [below=of r011] {}; 
		\node [nnlabel](nn3) [below=of n03,yshift=-6mm] {}; 
		\node [lnlabel](ln13) [below=of r013] {}; 
		\node [lnlabel](ln14) [below=of r014] {}; 
		%R- nodes
		\node [rnlabel](rn0) [below= of ln0] {}; %%The
		\node [word, left = of rn0, xshift=-1.4mm] {};
		\node [rnlabel](rn1) [below=of ln1] {}; %%cartridge
		%		\node [rnlabel](rn2) [below=of ln2] {}; %% was
		\node [rnlabel](rn3) [below=of ln3] {}; %% marked
		\node [rnlabel](rn4) [below=of ln4] {}; %% as 
		\node [rnlabel](rn5) [below=of ln5] {}; %% empty
		\node [rnlabel](rn6) [below=of ln6] {}; %% even
		%		\node [rnlabel](rn7) [below=of ln7] {}; %% with
		\node [srnlabel](rn8) [below=of ln8] {}; 
		\node [srnlabel](rn9) [below=of ln9] {}; 
		\node [rnlabel](rn10) [below=of ln10] {}; 
		\node [rnlabel](rn11) [below=of ln11] {}; 
		%		\node [rnlabel](rn12) [below=of ln12] {}; 
		\node [rnlabel](rn13) [below=of ln13] {}; 
		\node [rnlabel](rn14) [below=of ln14] {};

		%		\draw [chainLine] (lp0) to (lp1) ;
		%		\draw [chainLine] (lp1) to (np1) ;
		%		\draw [chainLine] (np1) to (rp3) ;
		%		\draw [chainLine] (rp3) to (rp4) ;
		%		\draw [chainLine] (rp4) to (ln5) ;
		%		\draw [chainLine] (ln5) to (ln6) ;
		%		\draw [chainLine] (ln6) to (nn2) ;
		%		\draw [chainLine] (nn2) to (rn8) ;
		%		\draw [chainLine] (rn8) to (rn9) ;
		%		\draw [chainLine] (rn9) to (rn10) ;
		%		\draw [chainLine] (rn10) to (lp11) ;
		%		\draw [chainLine] (lp11) to (np3) ;
		%		\draw [chainLine] (np3) to (rp13) ;
		%		\draw [chainLine] (rp13) to (rp14) ;
		
		\foreach \from/\to in {{lp0/lp1},{lp1/np1},{np1/rp3},{rp3/rp4},{ln5/ln6},{ln6/nn2},{nn2/rn8},{rn8/rn9},{lp10/lp11},{lp11/np3},{np3/rp13},{rp13/rp14}}
		\draw[->,blue]   (\from) to[out=0,in=180] (\to);
		\foreach \from/\to in {{rp4/ln5}}
		\draw[->,blue]   (\from) to[out=-15,in=145] (\to);
		\foreach \from/\to in {{rn9/lp10}}
		\draw[->,blue]   (\from) to[out=45,in=-145] (\to);

		% QT(s)
		%L+ nodes
		\node [slplabel](slp0) [below= of rn0, yshift = -4mm] {}; %%The
		\node [word, left = of slp0, xshift=-5.8mm,,yshift=-17mm] {{\textsc{QT(s)}}};
		\node [lplabel](slp1) [right=of slp0, xshift=2.5mm, yshift = -0mm] {}; 
		\node [lplabel](slp2) [right=of slp1, xshift=2.8mm] {}; 
		\node [lplabel](slp3) [right=of slp2, xshift=2.8mm] {}; 
		\node [lplabel](slp4) [right=of slp3,xshift=2.0mm] {}; 
		\node [lplabel](slp5) [right=of slp4,xshift=2.8mm] {}; 
		\node [lplabel](slp6) [right=of slp5, xshift=3mm] {}; %% even
		\node [lplabel](slp7) [right=of slp6, xshift=3mm] {}; %% with
		\node [lplabel](slp8) [right=of slp7, xshift=3.2mm] {}; 
		\node [lplabel](slp9) [right=of slp8, xshift=2.5mm] {}; 
		\node [slplabel](slp10) [right=of slp9, xshift=3mm] {}; 
		\node [lplabel](slp11) [right=of slp10, xshift=3mm] {}; 
		\node [lplabel](slp12) [right=of slp11,xshift=3.5mm] {}; 
		\node [lplabel](slp13) [right=of slp12,xshift=3.5mm] {}; 
		\node [lplabel](slp14) [right=of slp13,xshift=3.5mm] {}; 
		%R+ nodes
		\node [rpblabel](srp0) [below= of slp0] {}; %%The
		\node [word, left = of rp0, xshift=-2.4mm] {}; %{\textbf{\textsc{QT(s)}}};
		\node [rpblabel](srp1) [below=of slp1] {}; %%cartridge
		\node [rpblabel](srp2) [below=of slp2] {}; %% was
		\node [rpblabel](srp3) [below=of slp3] {}; %% marked
		\node [srpblabel](srp4) [below=of slp4] {}; %% as 
		\node [rpblabel](srp5) [below=of slp5] {}; %% empty
		\node [rpblabel](srp6) [below=of slp6] {}; %% even
		\node [rpblabel](srp7) [below=of slp7] {}; %% with
		\node [rpblabel](srp8) [below=of slp8] {}; 
		\node [rpblabel](srp9) [below=of slp9] {}; 
		\node [rpblabel](srp10) [below=of slp10] {}; 
		\node [rpblabel](srp11) [below=of slp11] {}; 
		\node [rpblabel](srp12) [below=of slp12] {}; 
		\node [rpblabel](srp13) [below=of slp13] {}; 
		\node [srpblabel](srp14) [below=of slp14] {}; 
		
		%L0 nodes
		\node [l0label](sl00) [below= of srp0] {}; %%The
		\node [word, left = of l00, xshift=-3.4mm]{};
		\node [l0label](sl01) [below=of srp1] {}; %%cartridge
		\node [l0label](sl02) [below=of srp2] {}; %% was
		\node [l0label](sl03) [below=of srp3] {}; %% marked
		\node [l0label](sl04) [below=of srp4] {}; %% as 
		\node [l0label](sl05) [below=of srp5] {}; %% empty
		\node [l0label](sl06) [below=of srp6] {}; %% even
		\node [l0label](sl07) [below=of srp7] {}; %% with
		\node [l0label](sl08) [below=of srp8] {}; 
		\node [l0label](sl09) [below=of srp9] {}; 
		\node [l0label](sl010) [below=of srp10] {}; 
		\node [l0label](sl011) [below=of srp11] {}; 
		\node [l0label](sl012) [below=of srp12] {}; 
		\node [l0label](sl013) [below=of srp13] {}; 
		\node [l0label](sl014) [below=of srp14] {}; 
		%R0 nodes
		\node [r0label](sr00) [below= of sl00] {}; %%The
		\node [word, left = of r00, xshift=-1.4mm]{};
		\node [r0label](sr01) [below=of sl01] {}; %%cartridge
		\node [r0label](sr02) [below=of sl02] {}; %% was
		\node [r0label](sr03) [below=of sl03] {}; %% marked
		\node [r0label](sr04) [below=of sl04] {}; %% as 
		\node [r0label](sr05) [below=of sl05] {}; %% empty
		\node [r0label](sr06) [below=of sl06] {}; %% even
		\node [r0label](sr07) [below=of sl07] {}; %% with
		\node [r0label](sr08) [below=of sl08] {}; 
		\node [r0label](sr09) [below=of sl09] {}; 
		\node [r0label](sr010) [below=of sl010] {}; 
		\node [r0label](sr011) [below=of sl011] {}; 
		\node [r0label](sr012) [below=of sl012] {}; 
		\node [r0label](sr013) [below=of sl013] {}; 
		\node [r0label](sr014) [below=of sl014] {}; 
		
		%L- nodes 
		\node [lnlabel](sln0) [below= of sr00] {}; %%The
		\node [word, left = of ln0, xshift=-1.4mm]{} ;
		\node [lnlabel](sln1) [below=of sr01] {}; %%cartridge
		\node [lnlabel](sln2) [below=of sr02] {}; %% was
		\node [lnlabel](sln3) [below=of sr03] {}; %% marked
		\node [lnlabel](sln4) [below=of sr04] {}; %% as 
		\node [slnlabel](sln5) [below=of sr05] {}; %% empty
		\node [lnlabel](sln6) [below=of sr06] {}; %% even
		\node [lnlabel](sln7) [below=of sr07] {}; %% with
		\node [lnlabel](sln8) [below=of sr08] {}; 
		\node [lnlabel](sln9) [below=of sr09] {}; 
		\node [lnlabel](sln10) [below=of sr010] {}; 
		\node [lnlabel](sln11) [below=of sr011] {}; 
		\node [lnlabel](sln12) [below=of sr012] {}; 
		\node [lnlabel](sln13) [below=of sr013] {}; 
		\node [lnlabel](sln14) [below=of sr014] {}; 
		%R- nodes
		\node [rnlabel](srn0) [below= of sln0] {}; %%The
		\node [word, left = of rn0, xshift=-1.4mm] {};
		\node [rnlabel](srn1) [below=of sln1] {}; %%cartridge
		\node [rnlabel](srn2) [below=of sln2] {}; %% was
		\node [rnlabel](srn3) [below=of sln3] {}; %% marked
		\node [rnlabel](srn4) [below=of sln4] {}; %% as 
		\node [rnlabel](srn5) [below=of sln5] {}; %% empty
		\node [rnlabel](srn6) [below=of sln6] {}; %% even
		\node [rnlabel](srn7) [below=of sln7] {}; %% with
		\node [rnlabel](srn8) [below=of sln8] {}; 
		\node [srnlabel](srn9) [below=of sln9] {}; 
		\node [rnlabel](srn10) [below=of sln10] {}; 
		\node [rnlabel](srn11) [below=of sln11] {}; 
		\node [rnlabel](srn12) [below=of sln12] {}; 
		\node [rnlabel](srn13) [below=of sln13] {}; 
		\node [rnlabel](srn14) [below=of sln14] {};

		%		\draw [chainLine] (slp0) to (srp4) ;
		%		\draw [chainLine] (srp4) to (sln5) ;
		%		\draw [chainLine] (sln5) to (srn10) ;
		%		\draw [chainLine] (srn10) to (slp11) ;
		%		\draw [chainLine] (slp11) to (srp14) ;
		\foreach \from/\to in {{slp0/srp4},{sln5/srn9},{slp10/srp14}}
		\draw[->,blue]   (\from) to[out=0,in=180] (\to);
		\foreach \from/\to in {{srp4/sln5}}
		\draw[->,blue]   (\from) to[out=-15,in=145] (\to);
		\foreach \from/\to in {{srn9/slp10}}
		\draw[->,blue]   (\from) to[out=45,in=-145] (\to);
		%		\draw [chainLine] (g5) to (g6) ;
		%		\draw [chainLine] (g6) to (g7) ;
		%		\draw [chainLine] (g7) to (g8) ;
		%		\draw [chainLine] (g8) to (g9) ;
		%		\draw [chainLine] (g9) to (g10) ;
		%		\draw [chainLine] (g10) to (g11) ;
		%		\draw [chainLine] (g11) to (g12) ;
		%		\draw [chainLine] (g12) to (g13) ;
		%		\draw [chainLine] (g13) to (g14) ;
		%		\draw [chainLine] (g14) to (g15) ;
		%		\draw [chainLine] (g15) to (g16) ;
		
		%QT(r)
		%L+ nodes
		\node [lplabel](rlp0) [below= of srn0, yshift=-4.0mm] {}; %%The
		\node [word, left = of rlp0, xshift=-1.60mm,yshift=-17mm] {{\textsc{QT(r)\textcolor{white}{(s)}}}}; %{\textbf{QT}};
		\node [slplabel](rlp1) [right=of rlp0, xshift=2.5mm] {}; 
		\node [snplabel](rnp1) [right=of rlp1, xshift=2.8mm, yshift=-3.5mm] {}; 
		\node [lplabel](rlp3) [right=of rlp1, xshift=15.8mm] {}; 
		\node [lplabel](rlp4) [right=of rlp3,xshift=2.0mm] {}; 
		\node [lplabel](rlp5) [right=of rlp4,xshift=2.8mm] {}; 
		\node [lplabel](rlp6) [right=of rlp5, xshift=3mm] {}; %% even
		\node [nplabel](rnp2) [right=of rlp6, xshift=3mm, yshift=-3.5mm] {}; %% with
		\node [lplabel](rlp8) [right=of rlp6, xshift=16.8mm] {}; 
		\node [lplabel](rlp9) [right=of rlp8, xshift=2.5mm] {}; 
		\node [lplabel](rlp10) [right=of rlp9, xshift=3mm] {}; 
		\node [lplabel](rlp11) [right=of rlp10, xshift=3mm] {}; 
		\node [snplabel](rnp3) [right=of rlp11,xshift=3.5mm,yshift=-3.5mm] {}; 
		\node [lplabel](rlp13) [right=of rlp11,xshift=17.5mm] {}; 
		\node [lplabel](rlp14) [right=of rlp13,xshift=3.5mm] {}; 
		%R+ nodes
		\node [rpblabel](rrp0) [below= of rlp0] {}; %%The
		\node [word, left = of rrp0, xshift=-2.4mm] {}; %{\textbf{\textsc{QT(s)}}};
		\node [rpblabel](rrp1) [below=of rlp1] {}; %%cartridge
		%		\node [rpblabel](rp2) [below=of lp2] {}; %% was
		\node [srpblabel](rrp3) [below=of rlp3] {}; %% marked
		\node [srpblabel](rrp4) [below=of rlp4] {}; %% as 
		\node [rpblabel](rrp5) [below=of rlp5] {}; %% empty
		\node [rpblabel](rrp6) [below=of rlp6] {}; %% even
		%		\node [rpblabel](rp7) [below=of lp7] {}; %% with
		\node [rpblabel](rrp8) [below=of rlp8] {}; 
		\node [rpblabel](rrp9) [below=of rlp9] {}; 
		\node [rpblabel](rrp10) [below=of rlp10] {}; 
		\node [rpblabel](rrp11) [below=of rlp11] {}; 
		%		\node [rpblabel](rp12) [below=of lp12] {}; 
		\node [srpblabel](rrp13) [below=of rlp13] {}; 
		\node [srpblabel](rrp14) [below=of rlp14] {}; 
		
		%L0 nodes
		\node [l0label](rl00) [below= of rrp0] {}; %%The
		\node [word, left = of l00, xshift=-3.4mm]{};
		\node [l0label](rl01) [below=of rrp1] {}; %%cartridge
		\node [n0label](rn01) [below=of rnp1, yshift=-7.0mm] {}; %% was
		\node [l0label](rl03) [below=of rrp3] {}; %% marked
		\node [l0label](rl04) [below=of rrp4] {}; %% as 
		\node [l0label](rl05) [below=of rrp5] {}; %% empty
		\node [l0label](rl06) [below=of rrp6] {}; %% even
		\node [n0label](rn02) [below=of rnp2, yshift=-7.0mm] {}; %% with
		\node [l0label](rl08) [below=of rrp8] {}; 
		\node [l0label](rl09) [below=of rrp9] {}; 
		\node [l0label](rl010) [below=of rrp10] {}; 
		\node [l0label](rl011) [below=of rrp11] {}; 
		\node [n0label](rn03) [below=of rnp3, yshift=-7.0mm] {}; 
		\node [l0label](rl013) [below=of rrp13] {}; 
		\node [l0label](rl014) [below=of rrp14] {}; 
		%R0 nodes
		\node [r0label](rr00) [below= of rl00] {}; %%The
		\node [word, left = of r00, xshift=-1.4mm]{};
		\node [r0label](rr01) [below=of rl01] {}; %%cartridge
		%		\node [r0label](r02) [below=of l02] {}; %% was
		\node [r0label](rr03) [below=of rl03] {}; %% marked
		\node [r0label](rr04) [below=of rl04] {}; %% as 
		\node [r0label](rr05) [below=of rl05] {}; %% empty
		\node [r0label](rr06) [below=of rl06] {}; %% even
		%		\node [r0label](r07) [below=of l07] {}; %% with
		\node [r0label](rr08) [below=of rl08] {}; 
		\node [r0label](rr09) [below=of rl09] {}; 
		\node [r0label](rr010) [below=of rl010] {}; 
		\node [r0label](rr011) [below=of rl011] {}; 
		%		\node [r0label](r012) [below=of l012] {}; 
		\node [r0label](rr013) [below=of rl013] {}; 
		\node [r0label](rr014) [below=of rl014] {}; 
		
		%L- nodes 
		\node [lnlabel](rln0) [below= of rr00] {}; %%The
		\node [word, left = of ln0, xshift=-1.4mm]{} ;
		\node [lnlabel](rln1) [below=of rr01] {}; %%cartridge
		\node [nnlabel](rnn1) [below=of rn01,yshift=-6mm] {}; %% was
		\node [lnlabel](rln3) [below=of rr03] {}; %% marked
		\node [lnlabel](rln4) [below=of rr04] {}; %% as 
		\node [slnlabel](rln5) [below=of rr05] {}; %% empty
		\node [slnlabel](rln6) [below=of rr06] {}; %% even
		\node [snnlabel](rnn2) [below=of rn02,yshift=-6mm] {}; %% with
		\node [lnlabel](rln8) [below=of rr08] {}; 
		\node [lnlabel](rln9) [below=of rr09] {}; 
		\node [lnlabel](rln10) [below=of rr010] {}; 
		\node [lnlabel](rln11) [below=of rr011] {}; 
		\node [nnlabel](rnn3) [below=of rn03,yshift=-6mm] {}; 
		\node [lnlabel](rln13) [below=of rr013] {}; 
		\node [lnlabel](rln14) [below=of rr014] {}; 
		%R- nodes
		\node [rnlabel](rrn0) [below= of rln0] {}; %%The
		\node [word, left = of rn0, xshift=-1.4mm] {};
		\node [rnlabel](rrn1) [below=of rln1] {}; %%cartridge
		%		\node [rnlabel](rn2) [below=of ln2] {}; %% was
		\node [rnlabel](rrn3) [below=of rln3] {}; %% marked
		\node [rnlabel](rrn4) [below=of rln4] {}; %% as 
		\node [rnlabel](rrn5) [below=of rln5] {}; %% empty
		\node [rnlabel](rrn6) [below=of rln6] {}; %% even
		%		\node [rnlabel](rn7) [below=of ln7] {}; %% with
		\node [srnlabel](rrn8) [below=of rln8] {}; 
		\node [srnlabel](rrn9) [below=of rln9] {}; 
		\node [rnlabel](rrn10) [below=of rln10] {}; 
		\node [rnlabel](rrn11) [below=of rln11] {}; 
		%		\node [rnlabel](rn12) [below=of ln12] {}; 
		\node [rnlabel](rrn13) [below=of rln13] {}; 
		\node [rnlabel](rrn14) [below=of rln14] {};

		%		\draw [chainLine] (lp0) to (lp1) ;
		%		\draw [chainLine] (lp1) to (np1) ;
		%		\draw [chainLine] (np1) to (rp3) ;
		%		\draw [chainLine] (rp3) to (rp4) ;
		%		\draw [chainLine] (rp4) to (ln5) ;
		%		\draw [chainLine] (ln5) to (ln6) ;
		%		\draw [chainLine] (ln6) to (nn2) ;
		%		\draw [chainLine] (nn2) to (rn8) ;
		%		\draw [chainLine] (rn8) to (rn9) ;
		%		\draw [chainLine] (rn9) to (rn10) ;
		%		\draw [chainLine] (rn10) to (lp11) ;
		%		\draw [chainLine] (lp11) to (np3) ;
		%		\draw [chainLine] (np3) to (rp13) ;
		%		\draw [chainLine] (rp13) to (rp14) ;
		
		\foreach \from/\to in {{rlp1/rnp1},{rnp1/rrp3},{rrp3/rrp4},{rln5/rln6},{rln6/rnn2},{rnn2/rrn8},{rrn8/rrn9},{rnp3/rrp13},{rrp13/rrp14}}
		\draw[->,blue]   (\from) to[out=0,in=180] (\to);
		\foreach \from/\to in {{rrp4/rln5}}
		\draw[->,blue]   (\from) to[out=-15,in=145] (\to);
		\foreach \from/\to in {{rrn9/rnp3}}
		\draw[->,blue]   (\from) to[out=0,in=180] (\to);
		
		%\node at (-0,0) {${T}$};
		\node[draw, text width = 1.229\textwidth] at (9.335,1.0) {\emph{There are} {\bf \textcolor{myred}{$\mathbf{22}$}} \emph{walnut trees} currently in the park . Park workers will plant walnut trees today . When the workers are finished there \emph{will be} {\bf \textcolor{myred}{$\mathbf{55}$}} \emph{walnut trees} in the \emph{park .} {\bf \textcolor{myred}{How}} \emph{many walnut} trees did the workers plant today ?};
		
		%	\draw[ ] (-0.6cm,0.5cm) rectangle node{\emph{There are} {\bf 22} \emph{walnut trees} currently in the park . Park workers will plant walnut trees today . When the workers are finished there \emph{will be} {\bf 55} \emph{walnut trees} in the \emph{park .} {\bf How} \emph{many walnut} trees did the workers plant today ?} (19.3cm,2.3cm);
		%  densely dotted
		\draw[ ] (-0.6cm,-0.3cm) rectangle node{} (19.3cm,0.3cm);
		
		\draw[ ] (-0.6cm,-0.8cm) rectangle node{} (19.3cm,-5cm);
		
		\draw[ ] (-0.6cm,-5.3cm) rectangle node{} (19.3cm,-9.5cm);
		
		\draw[ ] (-0.6cm,-9.8cm) rectangle node{} (19.3cm,-14.0cm);
		
		\draw[densely dotted] (-0.45cm,-0.95cm) rectangle node{} (5.55cm,-2.15cm);
		
		\draw[densely dotted ] (6.0cm,-3.65cm) rectangle node{} (12.25cm,-4.95cm);
		
		\draw[densely dotted ] (12.65cm,-0.95cm) rectangle node{} (19.00cm,-2.15cm);

		\draw[densely dotted] (-0.45cm,-5.45cm) rectangle node{} (5.55cm,-6.7cm);
		
		\draw[densely dotted ] (6.0cm,-8.15cm) rectangle node{} (12.25cm,-9.45cm);
		
		\draw[densely dotted ] (12.65cm,-5.45cm) rectangle node{} (19.00cm,-6.65cm);
		
		\draw[densely dotted] (0.9cm,-9.95cm) rectangle node{} (5.55cm,-11.20cm);
		
		\draw[densely dotted ] (6.0cm,-12.65cm) rectangle node{} (12.25cm,-13.95cm);
		
		\draw[densely dotted ] (15.40cm,-9.85cm) rectangle node{} (19.00cm,-11.20cm);

		\end{tikzpicture}

	}
	\vspace{-1mm}
	\caption{Illustrations of assumptions made by \textsc{QT}, {\textsc{QT(s)}} and \textsc{QT(r)}, with possible paths (selected nodes are highlighted) built for the token sequence $\boldsymbol{t}$ ($J$=$3$), consisting of words  from the original problem text $T$.}
	\label{fig:quantity_span_graph}
	\vspace{-5mm}
\end{figure*}
Solving arithmetic word problem can thus be casted as a sequence labeling problem where we assign every quantity appearing in the problem text a {sign} (in the form of a {\em tag}) from the set $\{\mathbb{+1},\mathbb{0},\mathbb{-1}\}$.
We further assume there exists a latent {\em quantity span} that needs to be learned -- a  sequence of words surrounding each quantity, based on which tagging decisions could be made.
 
We demonstrate through experiments on benchmark data that, despite its relatively simple assumptions involved, our novel sequence labeling approach is able to yield significantly better results than various state-of-the-art models.
To the best of our knowledge, this is the first work that tackles the problem from a sequence labeling perspective.
Our code is publicly available at \url{https://github.com/zoezou2015/quantity_tagger}.

\section{Our Approach}

\subsection{A Tagging Problem} 
\label{sec:tagger}
We define $\mathbf{Q}$ = $(q_1, q_2, \dots, q_i, x, q_{i+1}, \cdots q_m)$ (0$<$$i$$<$$m$, $m$$\geq$2 in arithmetic word problems) as an ordered quantity sequence for a problem text $T$, where $q_i \in \mathbf{Q}$ represents a \emph{constant quantity} appearing in $T$, and $x$ stands for the \emph{unknown quantity} assigned to the question sentence.
$\mathbf{Q}$ maintains the same order as the quantities appearing in $T$.
The goal is to construct a valid math equation $E$.
%our goal is to
% reveal the underlying meaning of the problem description and find the solution to $A$, with the help of equation $E$.
This research investigates such a problem by sequentially tagging each quantity $q\in\mathbf{Q}$ with the most likely \emph{sign} from set $\mathcal{S}=\{\mathbb{+1},\mathbb{0},\mathbb{-1}\}$,
%This work defines such the math word problems as a sequence tagging task.
%For each constant quantity $q_i \in q$ sequentially appearing in the problem text $t$ and the unknown quantity $x$, we consider the possible labels of them.
%Since we mainly focus on addition-subtraction problems, we define the sign set $\mathbf{S}$ containing three possible signs, namely $\{+,0,-\}$, 
where ``$\mathbb{+(-)1}$" means  a quantity is positively (negatively) related to the question, i.e., the sign of the quantity should be +(-) when forming part of the equation; ``$\mathbb{0}$" means a quantity is irrelevant to the question and should be ignored.
%The value of the unknown quantity $x$ is the solution we are looking for, which is thus always relevant.
%Therefore, we consider 2 possible signs $\{\mathbb{+1},\mathbb{-1}\}$ for $x$, while other quantities may be tagged with any of the 3 possible signs.

%With the sign tags, w
Given a specific prediction of the signs to the quantities, we can form an equation as follows:
%\begin{equation}
%\resizebox{0.6\hsize}{!}{$%
%\sum_{q_i \in \mathbf{Q}/\{x\}} s_iq_i + s_xx = 0$%
%}%
%\end{equation}
\begin{align}
\sum_{q_i \in \mathbf{Q}/\{x\}} s_iq_i + s_xx = 0
\end{align}
where $s_i \in \{\mathbb{+1},\mathbb{0},\mathbb{-1}\}$ is the sign for the $i$-th constant quantity $q_i$, and $s_x \in \{\mathbb{+1},\mathbb{-1}\}$ is the sign for $x$.
The solution can be easily obtained.

\subsection{Quantity Tagger}
\label{sec:linear}
%Our goal is to build an approach to find the optimal tag sequence that assigns each quantity in the text its corresponding sign.
%We present our approach -- the \emph{Quantity Tagger} (QT) model in this section.

Our primary assumption is that, for each quantity, there exists an implicit \emph{quantity span} that resides in the problem text and can convey relevant information useful for determining the signs of the quantities.
The quantity span of a quantity is essentially a contiguous token sequence from the problem text that consists of the quantity itself and some surrounding word tokens.
%Such quantity spans are not explicitly given in training data but need to be learned.

Formally, our model needs to learn how to sequentially assign each \emph{quantity} $q\in\mathbf{Q}$ its optimal {sign} $s\in\mathcal{S}$.
This is a sequence labeling problem \cite{lample2016neural,zou-19-joint}.
Common sequence labeling tasks, such as NER and POS tagging, mainly consider one sentence at a time, and tag each token in the sentence.
However, our tagging problem typically involves multiple sentences where relatively unimportant information may be potentially included.
For instance, the second sentence of Problem 2 in Figure \ref{fig:problem_definition}, ``\emph{Park workers will plant walnut trees today}" describes background knowledge of the problem, but such information may not be useful for solving problems, yet even obstructive.

% based on a  sequence of tokens  for $q$, which we call  a \emph{quantity span}.
%However, unlike many existing sequence labeling problems, our problem involves much longer text with multiple sentences.

%We assume that there exists a quantity span for each quantity $q\in \mathbf{Q}$.
%Specifically, the quantity span of a particular quantity $q$ is a consecutive sequence of tokens, consisting of the quantity $q$ as well as its surrounding tokens.

%Rather than simply adopting windows of fixed sizes when defining such spans, we assume there exist variable-sized  spans for quantities, where the exact sizes of such spans can be learned from data.
%Hence, rather than taking the whole problem text into account, we consider token windows.
For each quantity $q\in \mathbf{Q}$, we first consider a token window consisting of $q$ and $J-1$ surrounding tokens located immediately to the left and right of $q$.
%The window size is $2J-1$.
This gives us a window of word tokens in the size of $2J-1$.
Next, such token windows for all quantities in $\mathbf{Q}$ are merged to form a new token sequence, denoted as $\boldsymbol{t}$.
Note that $\boldsymbol{t}$ is formed by concatenating token subsequences taken from $T$ and is in the length of $n$ (1$\leq$ $n$$\leq$ $N$, where $N$ is the length of $T$).
We assume the quantity spans are defined over such a token sequence $\boldsymbol{t}$ (rather than $T$), which we believe convey most relevant information for determining the signs for the quantities.
%Token windows of all quantities are collected, denoted as $\mathbf{x}$, which is a subsequence of the problem text $T$.
%We thus take into account only a window of surrounding tokens in a fixed size $J$ for a certain quantity.
% where $J$ is the length of token sequences in the left/right part of such the window.
%Exemplified by the quantity 7341, we consider its surrounding window with $L=5$ as ``\emph{total of 7341 blood cells}".
%We thus have a subsequence tokens $\mathbf{y}$ of the original problem text where all quantities from $\mathbf{Q}$ and their surrounding tokens are included and the same order as in the original text is maintained.
%We further explicitly assume that every token in $\mathbf{T'}$ strictly belongs to one quantity span.
%In other words, there is no overlapping among quantity spans and all tokens in $\mathbf{x}$ are covered by quantity spans.
%Figure \ref{fig:quantity_span_graph} illustrated an example of how the new sequence $\mathbf{T'}$ is formed when $J=3$.
Exemplified by Problem 2 in Figure \ref{fig:problem_definition}, we show an example token sequence  $\boldsymbol{t}$ with $J=3$ in Figure \ref{fig:quantity_span_graph}.
%We further explicitly assume that all tokens in $\mathbf{x}$ are covered by quantity spans.
%The model is then built on the new token sequence $\mathbf{x}$.
%\begin{figure}[t!]
%	\centering
%	\resizebox{0.44\textwidth}{!}
%	{\begin{tikzpicture}[scale=0.6, every node/.style={scale=0.9}]
%		
%		%		\path [transform shape] (0,0) pic{sentimentspan2};
%		\pgfmathsetmacro{\SPAN}{2.55}
%		\pgfmathsetmacro{\OFFSET}{0.25}
%		\pgfmathsetmacro{\cx}{-4.0}
%		\pgfmathsetmacro{\cy}{1}
%		
%		\node at (-3,5) {${T}$};
%		\node[draw, text width = 0.85\linewidth,] at (4,4) {\emph{There are} {\bf 22} \emph{walnut trees} currently in the park . Park workers will plant walnut trees today . When the workers are finished there \emph{will be} {\bf 55} \emph{walnut trees} in the \emph{park .} {\bf How} \emph{many walnut} trees did the workers plant today ?};
% 
% 		\node at (-3,0) {$\mathbf{T}'$};
% 		\node[draw, text width = 0.85\linewidth] at (4,0) {There are {\bf 22} walnut trees will be {\bf 55} walnut trees park . {\bf How} many walnut};
% 		
% 		\coordinate (a) at (-3,4.5);
% 		\coordinate (b) at (-3,0.5);
% 		\draw[->, >=latex, blue!20!white, line width=8pt]   (a) to node[black]{$J=3$} (b) ;
%		
%		\end{tikzpicture}}
%	\caption{\small An example of forming new token sequence $\mathbf{T'}$ (below) from the problem text $T$ (above) when $J=3$.}
%	\label{fig:subsequence}
%\end{figure}

To capture quantity span information, we design 9 different labels with different semantics: $\mathcal{H}$=\{{$\mathbf{L}_\mathbf{+}$}, \textbf{$\mathbf{L}_\mathbf{0}$}, \textbf{$\mathbf{L}_\mathbf{-}$}; \textbf{$\mathbf{N}_\mathbf{+}$}, \textbf{$\mathbf{N}_\mathbf{0}$}, \textbf{$\mathbf{N}_\mathbf{-}$}; \textbf{$\mathbf{R}_\mathbf{+}$}, \textbf{$\mathbf{R}_\mathbf{0}$}, \textbf{$\mathbf{R}_\mathbf{-}$}\}.

\squishlist
	\item The $\mathbf{N}$ nodes are used to indicate that the current token is a quantity.
	\item The $\mathbf{L}$ ($\mathbf{R}$) nodes are used to indicate that the current token appears within a quantity span of a given quantity but to the left (right) of the quantity. 
\squishend
The subscripts ``$\mathbf{+}$'', ``$\mathbf{0}$'', and ``$\mathbf{-}$'' are used to denote the sign ($\mathbb{+1}$, $\mathbb{0}$ and $\mathbb{-1}$ respectively) associated with the quantities (and quantity spans).

All quantities are explicitly given in the problem text.
Therefore, the $\mathbf{N}$ node is used to tag a word token if and only if the  token represents a quantity.
Otherwise, $\mathbf{L}$ and $\mathbf{R}$ nodes are considered. 
Furthermore, the unknown quantity is always relevant to the problem.
We thus tag it with either $\mathbf{N_\mathbf{+}}$ or $\mathbf{N_\mathbf{-}}$,
 while three types of $\mathbf{N}$ nodes are for all constant quantities.
%Other tokens are associated with \textbf{F} and \textbf{N} nodes.
%Figure \ref{fig:latent_crf} demonstrates a real example of such a quantity span graph.
%We highlight the correct paths that our system predicted.
As illustrated in Figure \ref{fig:quantity_span_graph},
only one node from $\mathcal{H}$ will be selected at each position.
%At each position, only one tag will be selected 
%some relevant nodes will be selected based on the context of quantities and question intends.
Sequentially connecting all such nodes will form a single path that reveals information about quantity spans selected for all quantities.
%However, as quantity spans are latent variables, there exist multiple paths that all yield the same sign assignment to the quantities.
%The nodes selected at different positions will then be connected with each other, forming a \emph{quantity span graph} that shows the signs for quantities and their context scope information.
%From this graph, we can easily extract the equation which can be used to calculate answers.
%Figure \ref{fig:quantity_span_graph} (QT) illustrates a quantity span graph with possible spans for the running example.

%\colorbox{red}{The explanation of an example}
%\begin{table*}[th]
%	\centering
%	%	\resizebox{\textwidth}{!}
%	{\begin{tabular}{|l|l|l|l|l|l|}
%			\hline
%			Text & Value & Type        & Verb      & Nsubj                  \\ \hline
%			Jason has \textbf{43} blue and 16 red marbles. & 43     & blue, marbles     & has & Jason    \\ \hline
%			Jason has 43 blue and \textbf{16} red marbles. & 16  & red, marbles & has & Jason           \\ \hline
%			Tom has \textbf{24} blue marbles.  & 24  & blue, marbles & has &Tom            \\ \hline
%			\textbf{How} many blue marbles do they have in all? & x     & blue, marbles & have      & they     \\ \hline
%	\end{tabular}}
%	\caption{Example of quantities and their associated attributes}
%	\label{tab:attribute}
%\end{table*}
Following CRF \cite{lafferty2001conditional}, we formulate our method as a log-linear model with latent variables.
Formally, given the problem text $T$, let $\boldsymbol{t} = (t_1,t_2,\dots,t_{n})$ be a token sequence as defined above, %$\boldsymbol{y} = (y_1, y_2, \dots, y_{n})$ ($y_i \in \mathbf{S}$) 
$\boldsymbol{y}$ be the corresponding label sequence, and
%$\boldsymbol{h} = (h_1, h_2, \dots, h_{n})$ ($h_i \in \mathcal{H}$) 
$\boldsymbol{h}$ be a latent variable that provides specific quantity span information for the ($\boldsymbol{t,y}$) tuple, we define:
\begin{align}
%\resizebox{0.8\hsize}{!}{$%
p(\boldsymbol{y}|\boldsymbol{t}) = \frac{\sum_{\mathbf{h}}\exp(\mathbf{w}^T \mathbf{f}(\boldsymbol{t},\boldsymbol{y},\boldsymbol{h}))}{ \sum_{\boldsymbol{y'},\boldsymbol{h}'}\exp(\mathbf{w}^T \mathbf{f}(\boldsymbol{t},\boldsymbol{y'},\boldsymbol{h}'))}
%$%
%}
\end{align}
where $\mathbf{w}$ is the feature weight vector, i.e., model parameters, and $\mathbf{f}$ is the feature vector defined over the triple ($\boldsymbol{t,y,h}$), $\mathbf{f}(\boldsymbol{t},\boldsymbol{y},\boldsymbol{h})$ returns a list of discrete features (refer to supplementary materials).
%Features defined in $\mathbf{f}(\boldsymbol{t},\boldsymbol{y},\boldsymbol{h})$ can be broadly categorized into two groups: contextual and structural features.

During training, we would like to minimize the negative log-likelihood of the training set:
\begin{align}
&\mathcal{L}(\mathbf{w}) = \sum_{i}\log\sum_{\boldsymbol{y'},\boldsymbol{h'}}\exp{(\mathbf{w}_{f}^{T}\mathbf{f}(\boldsymbol{t}^{(i)},\boldsymbol{y'},\boldsymbol{h'}))} \notag \\
&-\sum_{i}\log\sum_{\boldsymbol{h}}\exp{(\mathbf{w}_{f}^{T}\mathbf{f}(\boldsymbol{t}^{(i)},\boldsymbol{y}^{(i)},\boldsymbol{h}))}
\end{align}
where the $(\boldsymbol{t}^{(i)},\boldsymbol{y}^{(i)})$ is the $i$-th training instance.
The standard gradient-based methods can be used to optimize the above objective, such as L-BFGS \cite{liu1989limited}.
Gradients of the above function is given by:
\begin{align}
&\frac{\partial\mathcal{L}(\mathbf{w})}{\partial w_{k}} = \sum_{i}\mathbf{E}_{p(\boldsymbol{y'},\boldsymbol{h}|\boldsymbol{t}^{(i)})}[f_k(\boldsymbol{t}^{(i)},\boldsymbol{y'},\boldsymbol{h})] \notag \\
&-\sum_{i}\mathbf{E}_{p(\boldsymbol{h}|\boldsymbol{t}^{(i)},\boldsymbol{y}^{(i)})}[f_k(\boldsymbol{t}^{(i)},\boldsymbol{y}^{(i)},\boldsymbol{h})]
\end{align}
where $\mathbf{E}_p[\cdot]$ is the expectation under distribution $p$.

We can construct a lattice representation on top of the nodes shown in Figure \ref{fig:quantity_span_graph}.
The representation compactly encodes exponentially many paths, where each path corresponds to one possible label sequence. 
%There exists a topological ordering amongst all nodes appearing in such a lattice representation.
%Thanks to this w
Note that there exists a topological ordering amongst all nodes.
This allows us to apply a generalized forward-backward algorithm to perform exact marginal inference so as to calculate both objective and expectation values efficiently \cite{li2017learning,zou2018crosslingual}.
The MAP inference procedure can be done analogously, which is called during the decoding time.

\subsection{Model Variants}
We further consider two variants of our model.
\noindent
\textbf{Semi-Markov Variant}: 
Our first variant, namely {\textsc{QT(s)}},  employs the semi-Markov assumption \cite{sarawagi2005semi}, where $\mathbf{N}$ nodes are removed.
%Under Semi-Markov assumption \cite{sarawagi2005semi}, we study the variant of Quantity Tagger in this section.
%Unlike the Quantity Tagger, the 
%At each position/token, there are two types of nodes \textbf{F} and \textbf{E} with three different signs $\{+1,0,-1\}$.
Different from QT which makes the first-order Markov assumption,   \textsc{QT(s)} assumes $\mathbf{L}$ and $\mathbf{R}$ nodes are used to indicate the left and right boundaries of a quantity span respectively.
Thus the model constructs edges (where non-Markovian features can be defined) by directly connecting the exactly first $\mathbf{L}$ and the last $\mathbf{R}$ nodes of a span.

\noindent
\textbf{Relaxed Variant}:
%\label{sec:relax}
One assumption made by \textsc{QT} is: each word in 
$\boldsymbol{t}$ strictly belongs to a certain quantity span.
The variant \textsc{QT(r)} relaxes such a constraint.
In this variant, some tokens in $\boldsymbol{t}$ may not belong to any quantity spans. 
Considering the example shown in Figure \ref{fig:quantity_span_graph}, the token ``\emph{There}" in $\boldsymbol{t}$ may not belong to any spans.

\section{Experiments}
\label{sec:experiment}
We conduct experiments on two datasets, AddSub \cite{hosseini2014learning}, consisting of 395 addition-subtraction problems in English, and AS\_CN with 1,049 addition-subtraction problems in Chinese \cite{wang2017deep}.
For all of our experiments, we use the L-BFGS algorithm \cite{liu1989limited} for learning model parameters with $\ell2$ regularization coefficient of 0.01.
To tune the hyperparameter $J$, we randomly select $80\%$ instances of the training set for training and the rest $20\%$ for development.
We tune $J$ on the development set.

\begin{table}[t]
	\centering
	\scalebox{0.9}
	{
		\resizebox{0.46\textwidth}{!}{
			\begin{tabular}{|l|c|c|}
				\hline
				Model     	& AddSub 		& AS\_CN    \\ \hline
				\hline
%				Majority   	&   37.08 &27.74  \\ \hline
				\citet{hosseini2014learning}   &77.70 & -  \\ \hline
				\citet{kushman2014learning}   &  64.00 & -  \\ \hline
				\citet{koncel2015parsing}  & 77.00 & -   \\ \hline
				\citet{roy2015solving}  & 78.00 & 47.57  \\ \hline
				\citet{zhou2015learn}       & 53.14 & 51.48  \\ \hline
				\citet{mitra2016learning}  & 86.07 & -  \\ \hline
				\citet{roy2017unit}  & 60.99 & 47.71 \\
				\hline
				\citet{wang2017deep} & - & 20.64 \\
				\hline
				\citet{wang2018mathdqn} & 78.50 & -   \\ \hline
				\hline
				\textsc{QT(fix)}& 87.73 & 53.19  \\ \hline
				QT       &  \textbf{90.79} & 58.72  \\ \hline
				\textsc{QT(s)}  & 87.30 & 54.81 \\ \hline
				\textsc{QT(r)}  & 88.69 & \textbf{59.10} \\ \hline
				\hline 
				\textsc{QT(-ef)}& 60.44 & 56.53  \\ \hline
				{\textsc{QT(s-ef)}}& 63.49 & 52.62  \\    \hline
				{\textsc{QT(r-ef)}}& 67.52 & 57.48 \\\hline
	\end{tabular}}}
	\vspace{-2mm}
	\caption{Accuracy (\%) on AddSub and AS\_CN. -\textsc{ef}: without external features.}
	\label{tab:acc_addsub}
	\vspace{-5mm}
\end{table}

\subsection{Analysis}
Following standard evaluation procedures used in previous works \cite{hosseini2014learning,mitra2016learning}, we conduct 3-fold cross validation on AddSub and AS\_CN, and report accuracies in Table \ref{tab:acc_addsub}.
We make comparisons with a list of recent works\footnote{Results on AS\_CN are obtained by running publicly released systems.} and two baselines.
%\footnote{We run code released by \newcite{zhou2015learn,roy2017unit} on AddSub. For AS\_CN, we run existing works \cite{zhou2015learn,roy2015solving,roy2017unit}. Note that systems of \cite{roy2015solving,roy2017unit} do not directly support Chinese corpus. We run these two systems on AS\_CN with its original settings for English corpus. \newcite{wang2017deep} did not release code. They offered help to run their system on AS\_CN. Figures for other systems were reported in their paper \cite{hosseini2014learning,kushman2014learning,koncel2015parsing,mitra2016learning}}
%The \emph{Majority} baseline tags all constant quantities as ``$\mathbb{+1}$" and all unknowns as ``$\mathbb{-1}$".
Another is \textsc{QT(fix)} where the quantity span for each quantity is a fixed-size token window.
All of our proposed models consistently outperform previous research efforts.
These figures confirm the capability of our approach to provide more promising solutions to addition-subtraction problems.
We do not require any additional annotations which can be expensive,
while annotations like variable-word alignments and formulas are necessary for works of \cite{kushman2014learning,mitra2016learning}.

To investigate the power of features extracted by external tools, such as ConceptNet \cite{liu2004conceptnet} and Stanford CoreNLP tool \cite{manning2014stanford}, we conduct additional experiments on the afore-mentioned datasets, where we call such features \emph{external features} (see supplementary material), indicated as ``\textsc{-ef}". 
It is expected that the performance drops because such features are necessary for capturing evidence across sentences.
Especially, for the AddSub dataset, it affects a lot.
As discussed before \cite{hosseini2014learning,mitra2016learning}, there exists lots of irrelevant information and information gaps in AddSub.
We thus can infer the external features support our approach to be capable of bridging information gaps and recognizing irrelevant information for solving arithmetic problems.
%Interestingly, the accuracy of the Chinese dataset AS\_CN slightly drops by about 2 points on average.
%This shows that such a corpus also benefits from external resources.
Poor performance shows challenges to solve such problems in Chinese.

\textbf{Which of our variants works the best?} 
We observe that models with variable-sized quantity spans, namely QT, \textsc{QT(s)} and \textsc{QT(r)}, generally perform better than \textsc{QT(fix)} where the quantity spans are fixed token windows.
This shows the effectiveness of introducing the quantity span as a latent variable.
QT obtains the highest average accuracy on the AddSub and \textsc{QT(r)} outperforms other two variants on the AS\_CN.

\begin{table}[tp]
	\centering
	\resizebox{0.48\textwidth}{!}
	{\begin{tabular}{|l|ccccc|ccccc|}
			\hline
			\multicolumn{1}{|c|}{\multirow{2}{*}{Model}} &
			\multicolumn{5}{c|}{ AddSub} &
			\multicolumn{5}{c|}{ AS\_CN} \\
			&  $A_{S.S.}$  & $A_{M.S.}$&$F_{\mathbb{+}}$&$F_{\mathbb{0}}$&$F_{\mathbb{-}}$ & $A_{S.S.}$  & $A_{M.S.}$&$F_{\mathbb{+}}$&$F_{\mathbb{0}}$&$F_{\mathbb{-}}$ \\
			\hline
			\hline
			%			ZDC \citeyearpar{zhou2015learn}       & 49.27 	&  84.28     & 56.05   & 51.79  \\ \hline
			%			ExpTree \citeyearpar{roy2015solving}  & 60.28 	& 61.13
			%			& \textbf{57.84}   & 37.23  \\ \hline
			%			UnitDep \citeyearpar{roy2017unit}	  & 60.28	& 64.23		& 57.02   & 37.07  \\ \hline
			%			\hline
			\textsc{QT}        &\textbf{89.5} & \textbf{97.3} &\textbf{96.0}&\textbf{86.4}&\textbf{96.5}  &56.9  &60.3 &85.5&62.2&85.0  \\ \hline
			\textsc{QT(s)}     &86.5 & 91.2 &95.0&82.8&95.6  &53.6  &56.3  &85.3&\textbf{62.9}&84.3 \\ \hline
			\textsc{QT(r)}     &87.5 & 92.6 &95.4&82.5&96.0 &\textbf{57.03}  &\textbf{60.9} &\textbf{86.5}&62.9&\textbf{85.6} \\ \hline
	\end{tabular}}
	\vspace{-2mm}
	\caption{Accuracies on two types of problems and $F1$ scores for three types of signs of quantities. $A_{S.S.}$: accuracy of single-step problems (\%) ; $A_{M.S.}$ accuracy of multi-step problems (\%) ; $F_{\mathbb{+(-/0)}}$: $F1$ score of sign ``$\mathbb{+1(-1/0)}$" (\%). }
	\label{tab:single-multi}
	\vspace{-5mm}
\end{table}

\textbf{How does our approach perform on different types of problems?}
%To understand our approaches better, 
We divide problems into two categories: single-step and multi-step problems.
The equation of a single-step problem contains at most two constant quantities tagged with either ``$\mathbb{+1}$" or ``$\mathbb{-1}"$, while the equation for a multi-step problem has more than two constant quantities with signs of ``$\mathbb{+1}"$ or ``$\mathbb{-1}$".
%For each fold, we train models on the other two folds and separately evaluate on such two types of problems taken from this fold.
%We collect results for every fold and take the average accuracy, as reported in Table \ref{tab:single-multi}.
%Following the same procedures, we calculate $F1$ scores for three signs (``$\mathbb{+1}$",``$\mathbb{-1}$" and ``$\mathbb{0}$"), as listed in Table \ref{tab:single-multi}.
We report accuracy and $F1$ score in Table \ref{tab:single-multi}. 
According to empirical results illustrated in Table \ref{tab:single-multi}, our approach is able to give more accurate answers to multi-step problems, while the accuracy of single-step problems is lower.
% which seem relatively easier than multi-step problems, though.
On the other hand, three models have similar patterns in terms of performance for three types of signs.
The $F1$ scores for signs of ``$\mathbb{+1}$" and ``$\mathbb{-1}$" are higher than scores of ``$\mathbb{0}$".
After examining outputs, we found that problem texts of single-step problems often contain more than two constant quantities,
among which only two of them are supposed to be labeled as ``$\mathbb{+1}$" or ``$\mathbb{-1}$" and the rest should be tagged as ``$\mathbb{0}$".
%This means we need to recognize those quantities with the sign of ``$\mathbb{0}$".
However, incorrectly labeling an irrelevant quantity with ``$\mathbb{+1}$" or ``$\mathbb{-1}$" leads to wrong solutions to single-step problems.
This also reveals that one main challenge for automatically solving arithmetic word problems is to recognize the irrelevant quantities.
Failures in identifying irrelevant information may due to implicit information of problem text or the external tool issues.

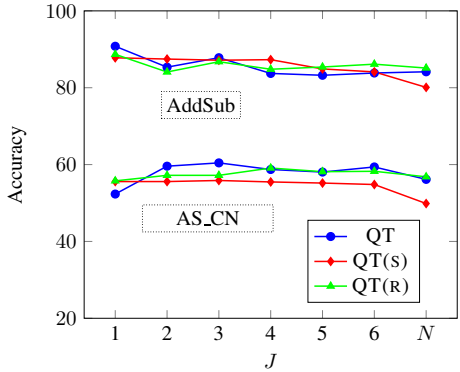
\begin{figure}[t]
	\centering
	\resizebox{0.78\linewidth}{!}
	{\begin{tikzpicture}
		\begin{axis}[xlabel= $J$, 
		ylabel=Accuracy,
		legend style={at={(0.9,0.32)},anchor=north east},
		symbolic x coords={1,2,3,4,5,6,$N$}, 
		xtick=data,ymin=20,ymax=100
		%x tick label style={rotate=0,anchor=east}
		]
		\addlegendentry{\textsc{QT}}
		\addplot[mark=*,thick,blue] coordinates {
			(1,90.79) (2,85.35) (3,87.81) (4,83.72) (5,83.24) (6,83.83) ($N$,84.15)
		};
		
		\addlegendentry{\textsc{QT(s)}}
		\addplot[mark=diamond*,thick,red] coordinates {
			(1,87.79) (2,87.46) (3,87.16) (4,87.30) (5,84.89) (6,84.12) ($N$,80.09)
		};
		
		\addlegendentry{\textsc{QT(r)}}
		\addplot[mark=triangle*,thick,green] coordinates {
			(1,88.69) (2,84.09) (3,86.77) (4,84.81) (5,85.39) (6,86.13) ($N$,85.12)
		};
		
		% 			\addlegendentry{\textsc{QT}}
		\addplot[mark=*,thick,blue] coordinates {
			(1,52.33) (2,59.58) (3,60.44) (4,58.72) (5,58.05) (6,59.38) ($N$,56.15)
		};
		
		% 			\addlegendentry{\textsc{QT(s)}}
		\addplot[mark=diamond*,thick,red] coordinates {
			(1,55.58) (2,55.58) (3,55.86) (4,55.48) (5,55.19) (6,54.81) ($N$,49.86)
		};
		
		% 			\addlegendentry{\textsc{QT(r)}}
		\addplot[mark=triangle*,thick,green] coordinates {
			(1,55.76) (2,57.20) (3,57.20) (4,59.10) (5,58.15) (6,58.25) ($N$,56.82)
		};
		%	\draw	(1.5cm,4.2cm) node{{\scriptsize AddSub}}
		%	(1.5cm,2cm) node{{\scriptsize AS\_CN}};
		\draw[densely dotted] (0.85cm,3.7cm) rectangle node{ AddSub} (2.3cm,4.2cm);
		\draw[densely dotted] (0.5cm,1.6cm) rectangle node{ AS\_CN} (2.9cm,2.1cm);
		%		\begin{scope}[shift={(4,4)}] 
		%		\draw (0.2,0.2) -- 
		%		plot[mark=*, mark options={fill=blue}] (0.25,0.2) -- (0.5,0.2) 
		%		node[right]{\textsc{QT}};
		%		\draw[yshift=\baselineskip] (0,0) -- 
		%		plot[mark=triangle*, mark options={fill=white}] (0.25,0) -- (0.5,0)
		%		node[right]{ciu};
		%		\draw[yshift=2\baselineskip] (0,0) -- 
		%		plot[mark=square*, mark options={fill=white}] (0.25,0) -- (0.5,0)
		%		node[right]{ciu + oscar};
		%		\draw[yshift=3\baselineskip] (0,0) -- 
		%		plot[mark=square*, mark options={fill=black}] (0.25,0) -- (0.5,0)
		%		node[right]{ciu + oscar extrapolated};
		%		\end{scope}
		\end{axis}
		\end{tikzpicture}}
	\vspace{-2mm}
	\caption{Effects of $J$ on three models (QT, \textsc{QT(s)} and \textsc{QT(r)}) evaluated on AddSub and AS\_CN. }
	\label{fig:line_chart}
	\vspace{-5mm}
\end{figure}

\textbf{Does $J$ really matter?} We further investigate the effects of $J$ on the three proposed models.
Figure \ref{fig:line_chart} plots how performance varies with $J$ ($J\in\{1,2,3,4,5,6,N\}$\footnote{All tokens in the problem text are considered as the selected token window for a quantity when $J=N$.}) on datasets AddSub (above) and AS\_CN (below).
%We can see that all three models have similar patterns.
On AddSub, three models have similar patterns that performance tends to be worse with a larger $J$.
As for the AS\_CN dataset, three models achieve relatively higher accuracies with $J\in\{2,3,6\}$ compared to other scenarios.
Interestingly, it seems that QT and \textsc{QT(r)} performs better than the semi-Markov variant \textsc{QT(s)}.
We tracked outputs from three models and found that \textsc{QT(s)} made more mistakes in predictions for unknown.
%Specifically the scenario with $J=N$ performs poorly which confirms our assumption taking token windows into account rather than the whole text is reasonable and effective.
The fact that models with $J=N$ perform do not perform well confirms our assumption that taking token windows into account rather than the whole text is reasonable and effective.

\textbf{Evaluation on different types of signs}: We investigate the capability of proposed approach to predict three types of signs ($\{\mathbb{+1,0,-1}\}$), as illustrated in Table \ref{tab:label}.
Three models have similar patterns on two datasets.
Predictions of ``$\mathbb{+1}$" and ``$\mathbb{-1}$" are more promising, compared to ``$\mathbb{0}$".
This reveals that one main challenge for automatically solving arithmetic word problems is to recognize the irrelevant information that should be labeled with ``$\mathbb{0}$".
Like what we discussed, failure on detecting irrelevant knowledge could be resulted from inevitably errors introduced by external resources and the lack of presence of crucial information in problem text.

\textbf{Error Analysis} The leading sources of errors can be categorized into three types: 1) The description of the problem is incomplete and implicit, which is challenging for machine to understand.
2) Failing in recognizing relevant quantities caused missing quantities or introducing irrelevant information.
3)
Incomplete information or errors from external tools, such as ConceptNet \cite{liu2004conceptnet} and Standford CoreNLP tool \cite{manning2014stanford}, are another source of errors leading to wrong predictions, which are inevitable.

\begin{table}[t]
	\centering
	\resizebox{0.45\textwidth}{!}
	{\begin{tabular}{|cc|ccc|ccc|}
			\hline
			\multicolumn{1}{|c}{\multirow{2}{*}{Model}} & &
			\multicolumn{3}{c|}{AddSub} &
			\multicolumn{3}{c|}{AS\_CN} 
			
			\\
			%\cline{2-3}
			&  &  $P.$\textcolor{white}{$\ast$} & $R.$\textcolor{white}{$\ast$} & $F.$\textcolor{white}{$\ast$} &  $P.$\textcolor{white}{$\ast$} & $R.$\textcolor{white}{$\ast$} & $F.$\textcolor{white}{$\ast$}  \\
			\hline
			\hline
			\multirow{3}{*}{QT}	&$\mathbb{+}$ &{\bf95.21}$\ast$  &{\bf96.70}$\ast$ &{\bf95.95}$\ast$ &81.86\textcolor{white}{$\ast$}  &89.54\textcolor{white}{$\ast$} &85.53\textcolor{white}{$\ast$}  \\
			&$\mathbb{0}$ &{\bf88.88}$\dagger$  &83.96\textcolor{white}{$\ast$}  &{\bf86.35}$\dagger$  &75.83\textcolor{white}{$\ast$}  &52.74\textcolor{white}{$\ast$}  &62.21\textcolor{white}{$\ast$}  \\ 
			&$\mathbb{-}$ &96.65\textcolor{white}{$\ast$}  &{\bf96.38}$\star$  &{\bf96.51}$\star$   &88.17\textcolor{white}{$\ast$}  &82.04\textcolor{white}{$\ast$}  &84.99\textcolor{white}{$\ast$}  \\\hline
			\multirow{3}{*}{\textsc{QT(s)}} & $\mathbb{+}$&93.97\textcolor{white}{$\ast$} &96.01\textcolor{white}{$\ast$} & 94.98\textcolor{white}{$\ast$}  &80.74\textcolor{white}{$\ast$}  &{\bf90.30}$\ast$  &85.26\textcolor{white}{$\ast$}  \\
			&$\mathbb{0}$	& 81.06\textcolor{white}{$\ast$} &{\bf84.50}$\dagger$   &82.75\textcolor{white}{$\ast$}  &75.00\textcolor{white}{$\ast$}  &{\bf54.22$\dagger$} & {\bf62.94}$\dagger$ \\
			&$\mathbb{-}$ &96.97\textcolor{white}{$\ast$} &94.18\textcolor{white}{$\ast$} & 95.55\textcolor{white}{$\ast$}  &{\bf88.66}$\star$  &80.25\textcolor{white}{$\ast$}  &84.25\textcolor{white}{$\ast$}  \\    \hline
			\multirow{3}{*}{\textsc{QT(r)}} & $\mathbb{+}$& 94.37\textcolor{white}{$\ast$}&96.42\textcolor{white}{$\ast$} & 95.38\textcolor{white}{$\ast$}  &{\bf83.79}$\ast$  &89.39\textcolor{white}{$\ast$}  &{\bf86.50}$\ast$  \\
			&$\mathbb{0}$	& 80.55\textcolor{white}{$\ast$} &{\bf84.50}$\dagger$   &82.48\textcolor{white}{$\ast$}  &{\bf78.02}$\dagger$  &52.72\textcolor{white}{$\ast$} & 62.92\textcolor{white}{$\ast$}    \\ 
			&$\mathbb{-}$ &{\bf97.48}$\star$ &94.65\textcolor{white}{$\ast$} & 96.04\textcolor{white}{$\ast$}  &86.67\textcolor{white}{$\ast$}  &{\bf84.61}$\star$  & {\bf85.63}$\star$ \\\hline
		\end{tabular}
	}
	\vspace{-2mm}
	\caption{Results for three types of signs for quantities predicted by three models. $P.$: Precision (\%), $R.$: Recall (\%), $F.$: $F1$ score (\%); Highest scores are in {\bf bold} and we use $\ast$, $\dagger$ and $\star$ to distinguish different sign types.}
	\label{tab:label}
	\vspace{-5mm}
\end{table}
\section{Conclusion and Future Work}

%This work proposes a novel model, \emph{Quantity Tagger}, to automatically solve a fundamental class of arithmetic word problems.
This work proposes the \emph{Quantity Tagger} that regards solving addition-subtraction problem as a sequence labeling task by introducing the quantity span for each quantity.
Despite its simplicity, it yields better performance.
In the future, we would also like to investigate better models that are capable to address general arithmetic word problems, including addition, subtraction, multiplication and division. 

\section*{Acknowledgments}
We would like to thank the three anonymous reviewers for their thoughtful and constructive comments.
We would also like to thank Yan Wang for his help on this work.
This work is supported by Singapore Ministry of Education Academic Research Fund (AcRF) Tier 2 Project MOE2017-T2-1-156, and is partially supported by SUTD project PIE-SGP-AI-2018-01.

\bibliography{addsub}
\bibliographystyle{acl_natbib}

\end{document}